\documentclass[11pt,a4paper]{article}
\usepackage[hyperref]{naaclhlt2019}
\usepackage{times}
\usepackage{latexsym}
\usepackage{graphicx}
\usepackage{caption}
\usepackage{booktabs}
\usepackage{array, makecell}
\usepackage{multirow}

\usepackage{pgfplotstable} %tikz plots data
\usepackage{pgfplots}
\pgfplotsset{compat=1.14}

\usepackage{url}

\aclfinalcopy % Uncomment this line for the final submission
\def\aclpaperid{1765} %  Enter the acl Paper ID here

\usepackage{amsmath,amsfonts,bm}

\def\eqref#1{equation~\ref{#1}}
\def\1{\bm{1}}

\def\mW{{\bm{W}}}
\def\mX{{\bm{X}}}

\DeclareMathAlphabet{\mathsfit}{\encodingdefault}{\sfdefault}{m}{sl}
\SetMathAlphabet{\mathsfit}{bold}{\encodingdefault}{\sfdefault}{bx}{n}

\newcommand{\R}{\mathbb{R}}

\renewcommand{\Re}{\mathbb{R}}
\newcommand*\samethanks[1][\value{footnote}]{\footnotemark[#1]}

\title{Cloze-driven Pretraining of Self-attention Networks}

\author{Alexei Baevski, Sergey Edunov\thanks{\hspace{0.06in}Equal contribution.} , Yinhan Liu\samethanks{}\hspace{0.05in}, Luke Zettlemoyer, Michael Auli \\
  Facebook AI Research \\
  Menlo Park, CA \and Seattle, WA \\
}

\date{}

\begin{document}
\maketitle
\begin{abstract}
We present a new approach for pretraining a bi-directional transformer model that provides significant performance gains across a variety of language understanding problems. 
Our model solves a cloze-style word reconstruction task, where each word is ablated and must be predicted given the rest of the text. 
Experiments demonstrate large performance gains on GLUE and new state of the art results on NER as well as constituency parsing benchmarks, consistent with the concurrently introduced BERT model.
We also present a detailed analysis of a number of factors that contribute to effective pretraining, including data domain and size, model capacity, and variations on the cloze objective.
\end{abstract}

\section{Introduction}

Language model pretraining has recently been shown to provide significant performance gains for a range of challenging language understanding problems~\citep{dai2015arxiv,peters2018acl,radford2018unsup}. 
However, existing work has either used unidirectional (left-to-right) language models (LMs)~\citep{radford2018unsup} or bi-directional (both left-to-right and right-to-left) LMs (BiLMs) where each direction is trained with an independent loss function~\cite{peters2018acl}. 
In this paper, we show that even larger performance gains are possible by jointly pretraining both directions of a large language-model-inspired self-attention cloze model. 
\begin{figure}[t]
\centering
\includegraphics[width=0.9\linewidth]{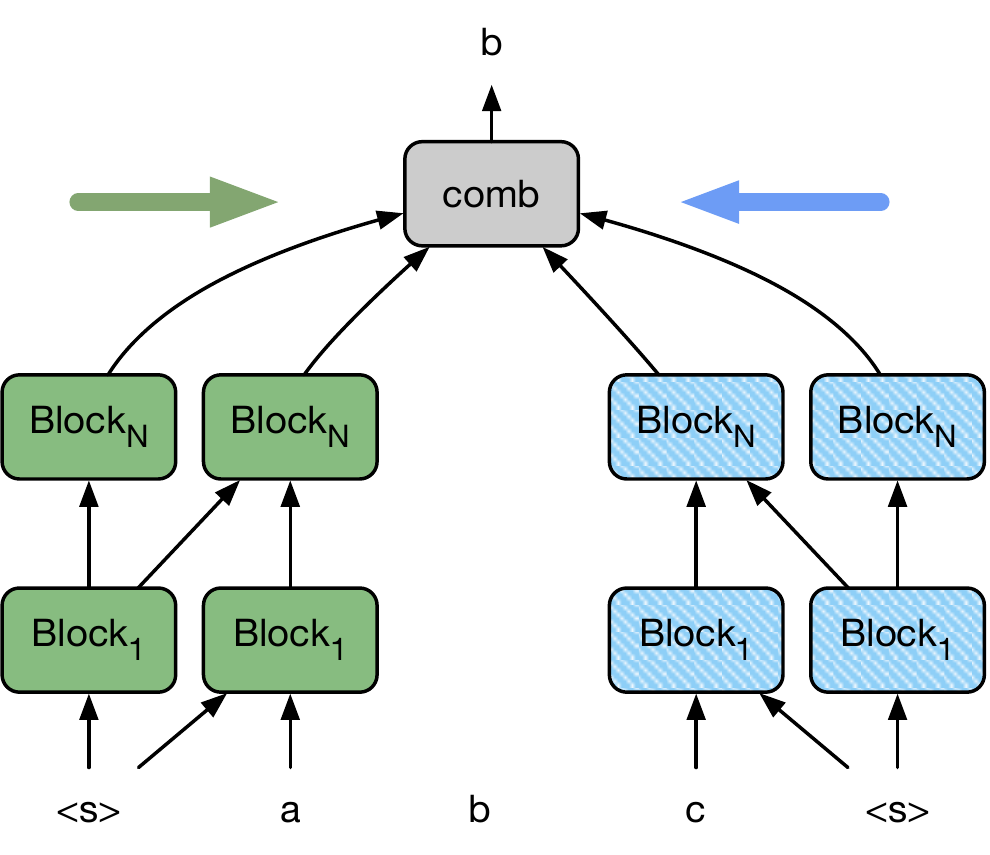}
\caption{Illustration of the model. Block$_i$ is a standard transformer decoder block. Green blocks operate left to right by masking future time-steps and blue blocks operate right to left. At the top, states are combined with a standard multi-head self-attention module whose output is fed to a classifier that predicts the center token.
}
\label{fig:model}
\end{figure}

Our bi-directional transformer architecture predicts \emph{every} token in the training data (Figure~\ref{fig:model}).
We achieve this by introducing a cloze-style training objective where the model must predict the center word given left-to-right and right-to-left context representations. 
Our model separately computes both forward and backward states with a masked self-attention architecture, that closely resembles a language model.
At the top of the network, the forward and backward states are combined to jointly predict the center word.
This approach allows us to consider both contexts when predicting words and to incur loss for every word in the training set, if the model does not assign it high likelihood. 

Experiments on the GLUE \citep{wang2018glue} benchmark show strong gains over the state of the art for each task, including a 9.1 point gain on RTE over \citet{radford2018unsup}. These improvements are consistent with, if slightly behind, those achieved by the concurrently developed BERT pretraining approach~\citep{devlin2018bert}, which we will discuss in more detail in the next section. We also show that it is possible to stack task-specific architectures for NER and constituency parsing on top of our pretrained representations, and achieve new state-of-the-art performance levels for both tasks. 
We also present extensive experimental analysis to better understand these results, showing that (1)  cross sentence pretraining is crucial for many tasks; (2) pre-training continues to improve performance with up to 18B tokens and would likely continue to improve with more data; and finally (3) our novel cloze-driven training regime is more effective than predicting left and right tokens separately.

\section{Related work}

There has been much recent work on learning sentence-specific representations for language understanding tasks.
\citet{mccann2017cove} learn contextualized word representations from a sequence to sequence translation task and uses the representations from the encoder network to improve a variety of language understanding tasks.
Subsequent work focused on language modeling pretraining which has been shown to be more effective and which does not require bilingual data \citep{zhang2018lm}.

Our work was inspired by ELMo \citep{peters2018acl} and the generative pretraining (GPT) approach of \citet{radford2018unsup}. 
ELMo introduces language models to pretrain word representations for downstream tasks including a novel mechanism to learn a combination of different layers in the language model that is most beneficial to the current task. 
GPT relies on a left to right language model and an added projection layer for each downstream task without a task-specific model.
Our approach mostly follows GPT, though we show that our model also works well with an ELMo module on NER and constituency parsing.

The concurrently introduced BERT model \citep{devlin2018bert} is a transformer encoder model that captures left and right context.
There is significant overlap between their work and ours but there are also significant differences: our model is a bi-directional transformer language model that predicts every single token in a sequence.
BERT is also a transformer encoder that has access to the entire input which makes it bi-directional but this choice requires a special training regime. 
In particular, they multi-task between predicting a subset of masked input tokens, similar to a denoising autoencoder, and a next sentence prediction task.
In comparison, we optimize a single loss function that requires the model to predict each token of an input sentence given all surrounding tokens.
We use all tokens as training targets and therefore extract learning signal from every single token in the sentence and not just a subset.

BERT tailors pretraining to capture dependencies between sentences via a next sentence prediction task as well as by constructing training examples of sentence-pairs with input markers that distinguish between tokens of the two sentences. 
Our model is trained similarly to a classical language model since we do not adapt the training examples to resemble the end task data and we do not solve a denoising task during training.

Finally, BERT as well as \citet{radford2018unsup} consider only a single data source to pretrain their models, either BooksCorpus \citep{radford2018unsup}, or BooksCorpus and additional Wikipedia data \citep{devlin2018bert}, whereas our study ablates the effect of various amounts of training data as well as different data sources.

\section{Two tower model}

Our cloze model represents a probability distribution  $p(t_i|t_1, \dots, t_{i-1}, t_{i+1}, \dots, t_n)$ for a sentence with $n$ tokens $t_1, \dots, t_n$.
There are two self-attentional towers each consisting of $N$ stacked blocks:
the \emph{forward} tower operates left-to-right and the \emph{backward} tower operates in the opposite direction.
To predict a token, we combine the representations of the two towers, as described in more detail below, taking care that neither representation contains information about the current target token.

The forward tower computes the representation $F^l_i$ for token $i$ at layer $l$ based on the forward representations of the previous layer $F^{l-1}_{\le i}$ via self-attention; the backward tower computes representation $B^l_i$ based on information from the opposite direction $B^{l-1}_{\ge i}$.
When examples of uneven length are batched, one of the towers may not have any context at the beginning. 
We deal with this issue by adding an extra zero state over which the self-attention mechanism can attend.

We pretrain on individual examples as they occur in the training corpora (\textsection\ref{sec:datasets}).
For News Crawl this is individual sentences while on Wikipedia, Bookcorpus, and Common Crawl examples are paragraph length.
Sentences are prepended and appended with sample boundary markers $<s>$.

\subsection{Block structure}
\label{sec:block}
The structure of the blocks follows most of the architectural choices described in \citet{vaswani2017transformer}.
Each block consists of two sub-blocks: the first is a multi-head self-attention module with $H=16$ heads for which we mask out any subsequent time-steps, depending on if we are dealing with the forward or backward tower.
The second sub-block is a feed-forward module (FFN) of the form $ReLU(\mW_1 \mX + b_1) \mW_2 + b_2$ where $\mW_1 \in \R^{e \times f}$, $\mW_1 \in \R^{f \times e}$.
Different to \citet{vaswani2017transformer} we apply layer normalization before the self-attention and FFN blocks instead of after, as we find it leads to more effective training.
Sub-blocks are surrounded by a residual connection \citep{he2015deep}.
Position is encoded via fixed sinusoidal position embeddings and we use a character CNN encoding of the input tokens for word-based models \citep{kim2016character}.
Input embeddings are shared between the two towers.

\subsection{Combination of representations}

The forward and backward representations computed by the two towers are combined to predict the ablated word.
To combine them we use a self-attention module which is followed by an FFN block (\textsection\ref{sec:block}).
The output of the FFN block is projected into $V$ classes representing the types in the vocabulary. 
When the model predicts token $i$, the input to the attention module are forward states $F^L_1 \dots F^L_{i-1}$ and backward states $B^L_{i+1} \dots B^:_n$ where $n$ is the length of the sequence and $L$ is the number of layers.
We implement this by masking $B^L_{\le i}$ and $F^L_{\ge i}$.
The attention query for token $i$ is a combination of $F^L_{i-1}$ and $B^L_{i+1}$.
For the base model we sum the two representations and for the larger models they are concatenated.
Keys and values are based on the forward and backward states fed to the attention module.
In summary, this module has access to information about the entire input surrounding the current target token.
During training, we predict every token in this way.
The output of this module is fed to an output classifier which predicts the center token. 
We use an adaptive softmax for the output classifier \citep{grave2017icml} for the word based models and regular softmax for the BPE based models.

While all states that contain information about the current target word are masked in the final self-attention block during training, we found it beneficial to disable this masking when fine tuning the pretrained model for downstream tasks. 
This is especially true for tasks that label each token, such as NER, as this allows the model to access the full context including the token itself.

\section{Fine-tuning}

\begin{figure}[t]
\centering
\includegraphics[width=0.8\linewidth]{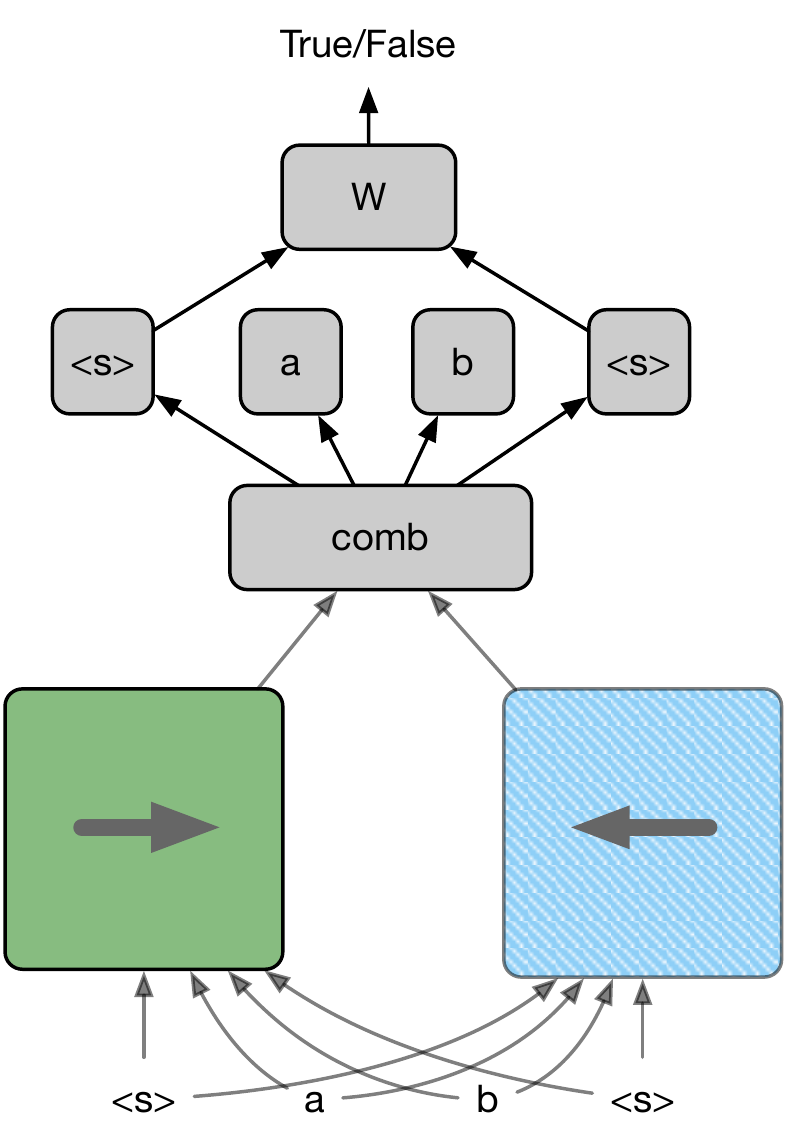}
\caption{Illustration of fine-tuning for a single-sentence task where the output of the first and last token is fed to a task-specific classifier (W).
Masking for the final combination layer (comb) is removed which results in representations based on all forward and backward states (cf. Figure~\ref{fig:model}).
}
\label{fig:finetune}
\end{figure}

We use the following approach to fine-tune the pretrained two tower model to specific downstream tasks (Figure~\ref{fig:finetune}).
\paragraph{Classification and regression tasks.}
For single sentence classification tasks, we consider the language model outputs for the boundary tokens $<s>$ which we add before the start and end of each sentence.
The outputs are of dimension $d=1024$ and we concatenate them to project to the number of classes $C$ in the downstream task with $W_{1} \in \Re^{C \times 2d}$ \citep{radford2018unsup};
we add a bias term $b \in \Re^C$ and initialize all weights as well as the bias to zero.
The output of the projection is softmax-normalized and the model is optimized with cross-entropy for classification tasks. 
Regression tasks such as the Semantic Textual Similarity benchmark (STS-B; Cer et al., 2017)\nocite{cer2017stsb} use $C=1$ and are trained with mean squared error.
For tasks involving sentence-pairs, we concatenate them and add a new separator token $<sep>$ between them. 
We add the output of this token to the final projection $W_{2} \in \Re^{C \times 3d}$.

\begin{table*}[t]
\centering
\begin{tabular}{lrrrrrcr}
\toprule
\bf Model & \bf Parameters & \bf Updates & \bf Blocks & \bf \thead{FFN\\ Dim} & \bf \thead{Attn Heads\\ (final layer)} & \bf \thead{Query formation\\ (final layer)} & \bf \thead{Train time\\ (days)} \\ \midrule
CNN Base & 177M & 600K & 6 & 4096 & 12 & Sum & 6 \\
CNN Large & 330M & 1M & 12 & 4096 & 32 & Concat & 10 \\
BPE Large & 370M & 1M & 12 & 4096 & 32 & Concat & 4.5 \\
\bottomrule
\end{tabular}
\caption{Hyper-parameters for our models. Parameter count excludes the (adaptive) softmax layer. Train time as measured on 128 Volta GPUs for the CNN models and 64 Volta GPUs for the BPE model.}
\label{tab:model_params}
\end{table*}

\paragraph{Structured prediction tasks.}
For named entity recognition and parsing we use task-specific architectures which we fine-tune together with the language model but with different learning rate.  
The architectures are detailed in the respective results sections.
The input to the architectures are the output representations of the pretrained language model.

\paragraph{No Masking.}
For fine-tuning, we found it beneficial to remove masking of the current token in the final layer that pools the output of the two towers. 
It is important to have access to information about the token to be classified for token level classification tasks such as NER but we also found this to perform better for sentence classification tasks. 
In practice, we completely disable masking in the combination layer so that it operates over all forward and backward states. However, disabling masking below the combination layer does not perform well.

\paragraph{Optimization.}
During fine-tuning we use larger learning rates for the new parameters, that is $W_1$, $W_2$, $b$ or the task-specific architecture, compared to the pretrained model.
For GLUE tasks, we do so by simply scaling the output of the language model before the $W_1$ and $W_2$ projections by a factor of 16. 
For structured prediction tasks, we explicitly use different learning rates for the pretrained model and the task-specific parameters.

We fine tune with the Adam optimizer \citep{kingma2015adam}. 
For GLUE tasks, we disable dropout in the language model and add 0.1 dropout between language model output and the final output projection; for structured prediction tasks, we use 0.3 at all levels (within the pretrained model, within the task-specific architecture, and on the weights connecting them).
In all settings, we use a batch size of 16 examples.
We use a cosine schedule to  linearly warm up the learning rate from 1e-07 to the target value over the first 10\% of training steps, and then anneal the learning rate to 1e-06, following the cosine curve for the remaining steps.
For GLUE tasks, we tuned the learning rate for each task and chose the best value over three settings: 1e-04, 5e-05 and 3e-05. 
For structured prediction tasks, we tuned on the pairs of learning rate, see the results section for details.  
For GLUE tasks, we train three seeds for each learning rate value for three epochs and choose the model after each epoch that performs best on the validation set. 
For structured prediction tasks, we train for up to 25 epochs and stop if the validation loss does not improve over the previous epoch.

\section{Experimental setup}

\subsection{Datasets for pretraining}
\label{sec:datasets}

We train the two tower model on several datasets.

\paragraph{Common Crawl.} We consider various subsets of Common Crawl which is web data.
We follow the same pre-processing as \citet{grave2018lrec} which is based on the May 2017 Common Crawl dump. 
This setup add 20 copies of English Wikipedia resulting in about 14\% of the final dataset to be Wikipedia. 
We subsample up to 18B tokens.
All experiments use Common Crawl subsampled to 9B tokens, except \textsection\ref{sec:data_exp}.

\paragraph{News Crawl.} We use up to 4.5B words of English news web data distributed as part of WMT 2018 \citep{bojar2018wmt}.

\paragraph{BooksCorpus + Wikipedia.} This is similar to the training data used by BERT which comprises the BooksCorpus \citep{zhu2015books} of about 800M words plus English Wikipedia data of 2.5B words.

\subsection{Pretraining hyper-parameters}

We adapt the transformer implementation available in the fairseq toolkit to our two tower architecture~\citep{ott2019fairseq}.
For hyper-parameter and optimization choices we mostly follow~\citet{baevski2018adp}.
Our experiments consider three model sizes shown in Table~\ref{tab:model_params}: 
There are two CNN input models in a base and large configuration as well as a Byte-Pair-Encoding based model (BPE; Sennrich et al., 2016)\nocite{sennrich:bpe:2016}.
The CNN models have unconstrained input vocabulary, and an output vocabulary limited to 1M most common types for the large model, and 700K most common types for the base model.
CNN models use an adaptive softmax in the output: the head band contains the 60K most frequent types with dimensionality 1024, followed by a 160K band with dimensionality 256. 
The remaining types have dimensionality 64; there are 480K types for the small model and 780K for the large model. 
The BPE model uses a vocabulary of 55K types and we share input and output embeddings in a flat softmax with dimension 1024 \citep{inan2016tying,press2016using}.
The BPE vocabulary was constructed by applying 30K merge operations over the training data, then applying the BPE code to the training data and retaining all types occurring at least three times.

Every setup uses model dimensionaltiy $d=1024$ with $H=16$ attention heads for all but the final attention layer. 
Model based on character inputs use character embedding size 128 and we apply six filters of size 1x128, 2x256, 3x384, 4x512, 5x512, 6x512 followed by a single highway layer. 
The models are trained with model and attention dropout rate of 0.1 and ReLU dropout rate of 0.05.

\begin{table*}[t]
\centering
\begin{tabular}{lrrrrrrrrr}
\toprule
& \thead{CoLA\\(mcc)} & \thead{SST-2\\(acc)} & \thead{MRPC\\(F1)} & \thead{STS-B\\(scc)} & \thead{QQP\\(F1)} & \thead{MNLI-(m/mm)\\(acc)} & \thead{QNLI\\(acc)} & \thead{RTE\\(acc)} & Avg \\
\midrule
OpenAI GPT     & 45.4 & 91.3 & 82.3 & 80.0 & 70.3 & 82.1/81.4 & 88.1 & 56.0 & 75.2\\
\midrule
CNN Base       & 53.1 & 93.6 & 81.3 & 82.2 & 70.5 & 82.5/82.2 & 89.5 & 64.6 & 77.7 \\
CNN Large      & 52.8 & 94.6 & 83.7 & 83.4 & 71.7 & 84.3/83.8 & 89.8 & 63.7 & 78.6 \\
BPE Large      & 51.8 & 94.0 & 83.0 & 84.2 & 70.6 & 82.9/82.2 & 89.3 & 65.1 & 78.1 \\
\midrule
GPT on STILTs  & 47.2 & 93.1 & 87.7 & 84.8 & 70.1 & 80.7/80.6 & 87.2 & 69.1 & 77.8 \\
BERT$_{BASE}$  & 52.1 & 93.5 & 88.9 & 85.8 & 71.2 & 84.6/83.4 & 90.1 & 66.4 & 79.6 \\
BERT$_{LARGE}$ & 60.5 & 94.9 & 89.3 & 86.5 & 72.1 & 86.7/85.9 & 91.1 & 70.1 & 81.9 \\
\bottomrule
\end{tabular}
\caption{Test results as per the GLUE evaluation server. The average column does not include the WNLI test set. mcc = Matthews correlation, acc = Accuracy, scc = Spearman correlation.
Concurrent work is shown below our results.}
\label{tab:glue}
\end{table*}

Different to \citet{vaswani2017transformer} we use Nesterov's accelerated gradient method \citep{sutskever2013icml} with a momentum of $0.99$ and we renormalize gradients if their norm exceeds $0.1$ \citep{pascanu2013difficulty}.
The learning rate is linearly warmed up from $10^{-7}$ to $1$ for 16K steps and then annealed using a cosine learning rate schedule with a single phase to 0.0001 \citep{cosine}.

We run experiments on DGX-1 machines with 8 NVIDIA V100 GPUs and machines are interconnected by Infiniband.
We also use the NCCL2 library and the torch.distributed package for inter-GPU communication.
We train models with 16-bit floating point precision, following~\citet{ott2018scaling}.
The BPE model trains much faster than the character CNN models (Table~\ref{tab:model_params}).

\section{Results}
     
\subsection{GLUE}
\label{sec:glue}

First, we conduct experiments on the general language understanding evaluation benchmark (GLUE; Wang et al., 2018)\nocite{wang2018glue} and present a short overview of the tasks. More information can be found in \citet{wang2018glue}. 
There are two single-sentence classification tasks: 
First, the Corpus of Linguistic Acceptability (CoLA; Warstadt et al., 2018)\nocite{warstadt2018cola} is a binary task to judge sentence grammaticality; evaluation is in terms of the Matthews correlation coefficient (mcc).
Second, the Stanford Sentiment Treebank (SST-2; Socher et al., 2013)\nocite{socher2013sst2} requires to judge if movie reviews have positive or negative sentiment; evaluation is in terms of accuracy (acc).

There are three tasks assessing sentence similarity:
The Microsoft Research Paragraph Corpus (MRPC; Dolan and Brockett, 2015)\nocite{dolan2005mrpc} and the Quora Question Pairs benchmark (QQP); we evaluate in terms of F1.
The Semantic Textual Similarity Benchmark (STS-B; Cer et al., 2017)\nocite{cer2017stsb} requires predicting a similarity score between 1 and 5 for a sentence pair; we report the Spearman correlation coefficient (scc).

Finally, there are four natural laguage inference tasks: the Multi-Genre Natural Language Inference (MNLI; Williams et al., 2018)\nocite{williams2017mnli}, the Stanford Question Answering Dataset (QNLI; Rajpurkar et al., 2016)\nocite{rajpurkar2016squad}, the Recognizing Textual Entailment (RTE; Dagan et al., 2006, Bar Haim et al., 2006, Ciampiccolo et al., 2007 Bentivogli et al., 2009)\nocite{dagan2006rte1,haim2006rte2,giampiccolo2007rte3,bentivogli2009rte}. 
We exclude the Winograd NLI task from our results similar to \citet{radford2018unsup,devlin2018bert} and report accuracy. 
For MNLI we report both matched (m) and mismatched (mm) accuracy on test.

We also report an average over the GLUE metrics. This figure is not comparable to the average on the official GLUE leaderboard since we exclude Winograd and do not report MRPC accuracy STS-B Pearson correlation as well as QQP accuracy.

Table~\ref{tab:glue} shows results for three configurations of our approach (cf. Table~\ref{tab:model_params}). 
The BPE model has more parameters than the CNN model but does not perform better in aggregate, however, it is faster to train.
All our models outperform the uni-directional transformer (OpenAI GPT) of \citet{radford2018unsup}, however, our model is about 50\% larger than their model.
We also show results for the concurrently introduced STILTs \citep{phang2018stilts} and BERT \citep{devlin2018bert}.
Our CNN base model performs as well as STILTs in aggregate, however, on some tasks involving sentence-pairs, STILTs performs much better (MRPC, RTE); there is a similar trend for BERT.

STILTs adds another fine-tuning step on another downstream task which is similar to the final task.
The technique is equally applicable to our approach.
Training examples for our model are  Common Crawl paragraphs of arbitrary length.
We expect that tailoring training examples for language model pretraining to the end tasks to significantly improve performance. 
For example, BERT trains on exactly two sentences while as we train on entire paragraphs.

\subsection{Structured Prediction}

We also evaluated performance on two structured predictions tasks, NER and constituency parsing. For both problems, we stacked task-specific architectures from recent work on top of our pretrained two tower models. We evaluate two ways of stacking: (1) ELMo-style, where the pretrained models are not fine-tuned but are linearly combined at different depths, and (2) with fine-tuning, where we set different learning rates for the task-specific layers but otherwise update all of the parameters during the task-specific training.

\begin{table}[t]
\centering
\begin{tabular}{lrrrrrcr}
\toprule
\bf Model & \bf dev F1 & \bf test F1 \\ \midrule
ELMo$_{BASE}$ &  95.7 & 92.2  \\ \midrule
CNN Large + ELMo & 96.4 & 93.2\\ 
CNN Large + fine-tune & 96.9 &93.5 \\ \midrule
BERT$_{BASE}$ & 96.4 & 92.4  \\
BERT$_{LARGE}$ & 96.6 & 92.8 \\
\bottomrule
\end{tabular}
\caption{ CoNLL-2003 Named Entity Recognition results. Test result was evaluated on parameter set with the best dev F1.}
\label{tab:NER_results}
\end{table}

\begin{table}[t]
\centering
\begin{tabular}{lrrrrrcr}
\toprule
\bf Model & \bf dev F1 & \bf test F1 \\ \midrule
ELMo$_{BASE}$ &  95.2 & 95.1  \\\midrule
CNN Large + ELMo & 95.1 & 95.2\\
CNN Large + fine-tune & 95.5 &95.6 \\
\bottomrule
\end{tabular}
\caption{Penn Treebank Constituency Parsing results. Test result was evaluated on parameter set with the best dev F1.}
\label{tab:parsing_results}
\end{table}

\begin{table*}
\centering
\begin{tabular}{lrrrrrrrrr}
\toprule
& \thead{CoLA\\(mcc)} & \thead{SST-2\\(acc)} & \thead{MRPC\\(F1)} & \thead{STS-B\\(scc)} & \thead{QQP\\(F1)} & \thead{MNLI-m\\(acc)} & \thead{QNLI\\(acc)} & \thead{RTE\\(acc)} & Avg \\
\midrule
cloze        & 55.1 & 92.9 & 88.3 & 88.3 & 87.2 & 82.3 & 86.5 & 66.4 & 80.9 \\
bilm         & 50.0 & 92.4 & 86.6 & 87.1 & 86.1 & 81.7 & 84.0 & 66.4 & 79.3 \\
cloze + bilm & 52.6 & 93.2 & 88.9 & 87.9 & 87.2 & 82.1 & 86.1 & 65.5 & 80.4 \\
\bottomrule
\end{tabular}
\caption{Different loss functions on the development sets of GLUE (cf. Table~\ref{tab:glue}). Results are based on the CNN base model (Table~\ref{tab:model_params})}
\label{tab:loss}
\end{table*}

\subsubsection{Named Entity Recognition}
\label{sec:ner}

We evaluated span-level F1 performance on the CoNLL 2003 Named Entity Recognition (NER) task, where spans of text must be segmented and labeled as Person, Organization, Location, or Miscellaneous.  We adopted the NER architecture in \citet{peters2018acl}, a biLSTM-CRF, with two minor modifications:  (1) instead of two layers of biLSTM, we only used one, and  (2) a linear projection layer was added between the token embedding and biLSTM layer. 
We did grid search on the pairs of learning rate, and found that projection-biLSTM-CRF with 1E-03 and pretrained language model with 1E-05 gave us the best result. 

Table~\ref{tab:NER_results} shows the results, with comparison to previous published ELMo$_{BASE}$ results \citep{peters2018acl} and the BERT models. Both of our stacking methods outperform the previous state of the art, but fine tuning gives the biggest gain.

\subsubsection{Constituency Parsing}
\label{sec:parsing}
We also report parseval F1 for Penn Treebank constituency parsing. We adopted the current state-of-the-art architecture~\cite{kitaev2018acl}. We again used grid search for learning rates and number of layers in parsing encoder, and used 8E-04 for language model finetuning, 8E-03 for the parsing model parameters, and two layers for encoder. 

Table~\ref{tab:parsing_results} shows the results. Here, fine tuning is required to achieve gains over the previous state of the art, which used ELMo embeddings. 

\subsection{Objective functions for pretraining}
\label{sec:losses}

The two-tower model is trained to predict the current token given representations of the entire left and right context (cloze).
Next we compare this choice to two alternatives:
First, \citet{peters2018acl} train two language models operating left-to-right and right-to-left to predict the next word for each respective direction.
We change the two-tower model to predict the next word using the individual towers only and remove the combination module on top of the two towers (bilm); however, we continue to jointly train the two towers.

Second, we combine the cloze loss with the bilm loss to obtain a triplet loss which trains the model to predict the current word given both left and right context, as well as just right or left context.
The latter is much harder than the cloze loss since less context is available and therefore gradients for the bilm loss are much larger: the cloze model achieves perplexity of about 4 while as for the bilm it is 27-30, depending on the direction.
This results in the bilm loss dominating the triplet loss and we found that scaling the bilm term by a factor of $0.15$ results in better performance.

Table~\ref{tab:loss} shows that the cloze loss performs significantly better than the bilm loss and that combining the two loss types does not improve over the cloze loss by itself.
We conjecture that individual left and right context prediction tasks are too different from center word prediction and that their learning signals are not complementary enough.

\subsection{Domain and amount of training data}
\label{sec:data_exp}

\begin{figure}[t]
\begin{center}
\resizebox {1\columnwidth}{!}{
\begin{tikzpicture}
\begin{axis}[
  xmode=log,
  xlabel=Train data tokens,
  ylabel=Avg. GLUE score,
  xtick={562,1125,2250,4500,9000,18000},
  xticklabels={562M,1.1B,2.25B,4.5B,9B,18B},
  ymax=81.5,
  style={thick},
  legend pos=north west,
  grid=both]
\addplot table [y=ccrawl, x=data]{data/ccrawl.dat};
\addlegendentry{Average GLUE score}
\end{axis}
\end{tikzpicture}
}
\caption{Average GLUE score with different amounts of Common Crawl data for pretraining.
\label{fig:data}}
\end{center}
\end{figure}
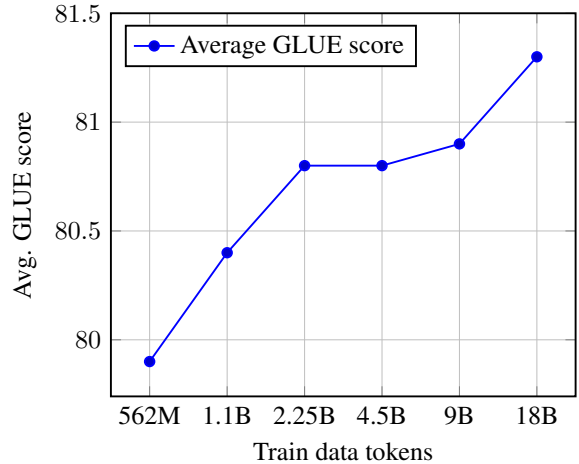

\begin{table*}[t!]
\centering
\begin{tabular}{lrrrrrrrrrr}
\toprule
& \thead{train data\\(M tok)} & \thead{CoLA\\(mcc)} & \thead{SST-2\\(acc)} & \thead{MRPC\\(F1)} & \thead{STS-B\\(scc)} & \thead{QQP\\(F1)} & \thead{MNLI-m\\(acc)} & \thead{QNLI\\(acc)} & \thead{RTE\\(acc)} & Avg \\
\midrule
\multirow{6}{*}{\shortstack{ccrawl}} 
& 562 & 52.5 & 92.9 & 88.2 & 88.3 & 87.1 & 81.7 & 85.7 & 63.3 & 79.9 \\
& 1125 & 55.5 & 93.1 & 86.1 & 88.4 & 87.1 & 81.9 & 85.7 & 65.2 & 80.4 \\
& 2250 & 55.4 & 92.4 & 87.7 & 88.4 & 87.2 & 82.2 & 86.2 & 66.9 & 80.8 \\
& 4500 & 56.6 & 93.0 & 87.3 & 88.6 & 87.0 & 82.0 & 86.2 & 65.7 & 80.8 \\
& 9000 & 55.1 & 92.9 & 88.3 & 88.3 & 87.2 & 82.3 & 86.5 & 66.4 & 80.9 \\
& 18000 & 56.3 & 93.1 & 88.0 & 88.8 & 87.2 & 82.3 & 86.3 & 68.4 & 81.3 \\
\midrule
\multirow{4}{*}{\shortstack{news\\crawl}} 
& 562 & 50.9 & 92.8 & 81.4 & 78.2 & 84.9 & 79.1 & 82.0 & 55.7 & 75.6 \\
& 1125 & 51.4 & 93.0 & 83.0 & 82.3 & 85.2 & 79.7 & 82.8 & 53.9 & 76.4 \\
& 2250 & 54.8 & 92.9 & 83.5 & 82.8 & 85.4 & 80.4 & 82.4 & 54.8 & 77.1 \\
& 4500 & 53.9 & 93.6 & 83.8 & 83.1 & 85.5 & 80.4 & 83.6 & 54.2 & 77.3 \\
\midrule
BWiki - sent & 3300 & 53.5 & 91.6 & 86.4 & 86.2 & 86.9 & 82.3 & 86.9 & 63.8 & 79.7 \\
BWiki - blck & 3300 & 50.6 & 91.9 & 86.4 & 87.1 & 86.8 & 81.9 & 86.2 & 60.4 & 78.9 \\
\bottomrule
\end{tabular}
\caption{Effect of different domains and amount of data for pretraining on the on the development sets of GLUE (cf. Table~\ref{tab:glue}). Results are based on the CNN base model (Table~\ref{tab:model_params}).
}
\label{tab:data}
\end{table*}

Next we investigate how much pretraining benefits from larger training corpora and how the domain of the data influences end-task performance.

Figure~\ref{fig:data} shows that more training data can significantly increase accuracy. 
We train all models with the exact same hyper-parameter settings on Common Crawl data using the CNN base architecture for 600K updates.
We train on up to 18B Common Crawl tokens and the results suggest that more training data is likely to further increase performance.

Table~\ref{tab:data} shows a breakdown into individual GLUE tasks.
For pretraining on Common Crawl, CoLA and RTE benefit most from additional training data.
The same table also shows results for News Crawl which contains newswire data.
This data generally performs less well than Common Crawl, even on MRPC which is newswire.
A likely reason is that News Crawl examples are \emph{individual sentences} of 23 words on average which compares to several sentences or 50 words on average for Common Crawl.
Mutli-sentence training examples are more effective for end-tasks based on sentence pairs, e.g., there is a 14 point accuracy gap on RTE between News Crawl and Common Crawl with 4.5B tokens.
More News Crawl data is most beneficial for CoLA and STS-B. 

We also experiment with BooksCorpus \citep{zhu2015books} as well as English Wikipedia, similar to \citet{devlin2018bert}. 
Examples in BooksCorpus are a mix of individual sentences and paragraphs; examples are on average 36 tokens.
Wikipedia examples are longer paragraphs of 66 words on average.
To reduce the effect of training on examples of different lengths, we adopted the following strategy: we concatenate all training examples into a single string and then crop blocks of $512$ consecutive tokens from this string. 
We train on a batch of these blocks (BWiki - blck).
It turns out that this strategy did not work better compared to our existing strategy of simply using the data as is (BWiki - sent).
BooksCorpus and Wikipedia performs very well on QNLI and MNLI but less well on other tasks.

In summary, more data for pretraining improves performance, keeping everything else equal. 
Also pretraining on corpora that retains paragraph structure performs better than individual sentences.

\section{Conclusion}

We presented a pretraining architecture based on a bi-directional transformer model that predicts every token in the training data.
The model is trained with a cloze-style objective and predicts the center word given all left and right context.

Results on the GLUE benchmark show large gains over \citet{radford2018unsup} for each task, while experiments with model stacking set new state of the art performance levels for parsing and named entity recognition. We also did extensive experimental analysis to better understand these results, showing that (1)  cross sentence pretraining is crucial for many tasks; (2) pre-training continues to improve performance up to 18B tokens and would likely continue to improve with more data; and finally (3) our novel cloze-driven training regime is more effective than predicting left and right tokens separately.

In future work, we will investigate variations of our architecture. In particular, we had initial success sharing the parameters of the two towers which allows training much deeper models without increasing the parameter count.

\bibliography{master}
\bibliographystyle{acl_natbib}

\end{document}